\documentclass{article}
\usepackage{spconf,amsmath,graphicx}
\usepackage{wrapfig}
\usepackage{hyperref}
\usepackage{multirow}
\usepackage{subfig}
\usepackage{booktabs}
\usepackage{xcolor}

\title{ERSAM: Neural Architecture Search For\\\underline{E}nergy-Efficient and \underline{R}eal-Time \underline{S}ocial \underline{A}mbiance \underline{M}easurement}

\name{Chaojian Li$^{*,1}$\thanks{$^{*}$Equal contribution.}, Wenwan Chen$^{*,2}$, Jiayi Yuan$^{*,2}$, Yingyan (Celine) Lin$^{1}$, and Ashutosh Sabharwal$^{2}$\thanks{\textcolor{black}{We would like to acknowledge the funding support from the NSF SCH, Expedition, and RTML programs (Award ID: 1838873, 1730574, and 1937592) for this project.}}}
\address{ $^{1}$Georgia Institute of Technology, Atlanta, GA \\
$^{2}$Rice University, Houston, TX}

\begin{document}
\ninept

\maketitle
\begin{abstract}

Social ambiance describes the context in which social interactions happen, and can be measured using speech audio by counting the number of concurrent speakers. This measurement has enabled various mental health tracking and human-centric IoT applications. While on-device Socal Ambiance Measure (SAM) is highly desirable to ensure user privacy and thus facilitate wide adoption of the aforementioned applications, the required computational complexity of state-of-the-art deep neural networks (DNNs) powered SAM solutions stands at odds with the often constrained resources on mobile devices. Furthermore, only limited labeled data is available or practical when it comes to SAM under clinical settings due to various privacy constraints and the required human effort, further challenging the achievable accuracy of on-device SAM solutions. 
To this end, we propose a dedicated neural architecture search framework for \textbf{E}nergy-efficient and \textbf{R}eal-time \textbf{SAM} (ERSAM). Specifically, our ERSAM framework can automatically search for DNNs that push forward the achievable accuracy vs. hardware efficiency frontier of mobile SAM solutions. For example, ERSAM-delivered DNNs only consume 40 mW $\cdot$ 12 h energy and 0.05~seconds processing latency for a 5~seconds audio segment on a Pixel~3 phone, while only achieving an error rate of 14.3\%  on a social ambiance dataset generated by LibriSpeech. We can expect that our ERSAM framework can pave the way for ubiquitous on-device SAM solutions which are in growing demand. 

\end{abstract}

\section{Introduction}
\label{sec:intro}
Social ambiance, which describes the context of an environment where social interactions are happening, can be measured via speech audio~\cite{chen2021privacy,wang2014local}. Among the widely used social ambiance measure (SAM) solutions, counting the number of concurrent speakers around an individual has been verified to be associated with their mental health symptoms (e.g., depression and psychotic disorders)~\cite{chen2021privacy}.
While deep neural networks (DNNs) have enabled accurate SAM in mental health tracking by counting concurrent speakers~\cite{chen2021privacy}, it is still challenging to achieve continuous, real-time, and on-device DNN-based SAM, which is highly desirable for preserving user privacy and thus ensuring wide adoption of the aforementioned mental health tracking. 
\underline{First}, powerful DNNs are often complex while mobile/wearable devices are very limited in both computational and memory resources, e.g., deploying the wav2vec2 DNN~\cite{baevski2020wav2vec} on a Pixel 3 phone~\cite{pixel3} for speech recognition requires 2200 mW~\cite{pytorch_mobile_wav2vec2} power consumption vs. only 700 mW~\cite{zhidkov2018smartphone} allowed by the Pixel 3 phone~\cite{pixel3}.   
\underline{Second}, training DNNs usually requires a large amount of labeled data to achieve a satisfactory task accuracy, whereas SAM users expect DNNs to be trained locally on a device with limited data (e.g., less than 8 hours of training data~\cite{xu2020lrspeech}), due to both the privacy concern and prohibitive human effort needed to collect clinical SAM audio data.

\begin{figure}[t]
    \centering
    \includegraphics[width=1.0\linewidth]{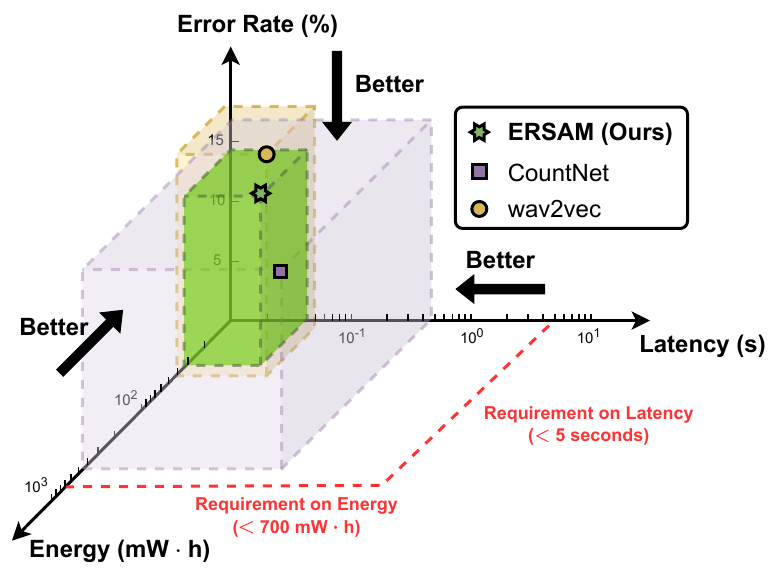}
    \vspace{-2em}
   \caption{DNNs searched by our ERSAM outperform state-of-the-art DNN-based SAM: achieving the best accuracy vs. hardware efficiency trade-offs while meeting the requirements of being energy-efficient and achieving real-time latency. Here a smaller volume of each data point's 3D cuboid corresponds to a better accuracy vs. hardware efficiency trade-off.}
    \label{fig:performance_overview}
    \vspace{-2em}
\end{figure}

To close the aforementioned gap between the required computational complexity for DNN-based SAM and the limited resources on mobile/wearable devices, it is critical to develop compact DNNs that ensure \underline{(i)} real-time latency (e.g., \textbf{$\leq$ 5 seconds of processing time for an audio segment of 5 seconds}~\cite{chen2021privacy}) for timely usage of the generated SAM information (e.g., real-time intervention),
\underline{(ii)} constrained energy (\textbf{$\leq$ 700mW $\cdot$ 12h~\cite{zhidkov2018smartphone}}), and \underline{(iii)} an acceptable SAM accuracy under the aforementioned low-resource settings.
Thanks to the recent success of both \underline{(i)} \textbf{H}ard\textbf{w}are-aware \textbf{N}eural \textbf{A}rchitecture \textbf{S}earch (HW-NAS)~\cite{wu2019fbnet} in developing DNN models with optimal accuracy vs. hardware efficiency trade-offs~\cite{wu2019fbnet,cai2019once,yu2020bignas} and \underline{(ii)} knowledge distillation in leveraging pre-trained giant models to boost the achievable accuracy of compact DNNs ~\cite{hinton2015distilling,gou2021knowledge}, it is natural to consider them for continuous, real-time, and on-device DNN-based SAM solutions. 
However, existing HW-NAS and knowledge distillation techniques are not directly applicable for meeting the above requirements for the following reasons: \underline{(i)} prior HW-NAS works mainly focus on computer vision or natural language processing tasks~\cite{wu2019fbnet,cai2019once}, where the search space of existing HW-NAS works might not be optimal or even feasible for DNNs dedicated to SAM due to their differences from SAM in terms of 1) input modalities (e.g., images vs. audio), 2) commonly used operators (e.g., 2D convolution vs. 1D convolution), and 3) dataset size (e.g., 150 GB ImageNet~\cite{deng2009imagenet} with $>$ 1 million images vs. 1 GB dataset with only 8 hours of audio);
and \underline{(ii)} the large cost of querying the giant models required by vanilla knowledge distillation techniques 
makes knowledge distillation impractical to be used on a local edge/mobile device.

Our key contribution is a new framework, called neural architecture search for Energy-efficient and Real-time SAM (ERSAM), which achieves continuous, real-time, and on-device DNN-based SAM, as elaborated below.

\begin{itemize}
    \item We develop and validate a dedicated HW-NAS and knowledge distillation framework called ERSAM to develop DNN-based SAM solutions toward meeting SAM's real-world application driven requirements. 
    \item ERSAM Enabler 1: The proposed ERSAM integrates a hardware-aware search space dedicated to SAM by leveraging the cost profiling observations from state-of-the-art DNN-based SAM models on mobile phones. 
    \item ERSAM Enabler 2: We propose an efficient knowledge distillation scheme to be embedded into both the search and training processes of our ERSAM framework's HW-NAS engine, which enables the feasibility of developing compact yet effective DNNs for SAM on the local edge by only making necessary queries to the giant teacher model.
    \item ERSAM performance: As shown in Fig. \ref{fig:performance_overview}, our ERSAM framework's delivered DNNs fulfill all the above requirements, achieving \underline{(i)} \textbf{0.05 seconds ($\leq$ 5 seconds~\cite{chen2021privacy})} latency for an audio segment of 5 seconds, \underline{(ii)} \textbf{40 mW $\cdot$ 12 h ($\leq$ 700 mW $\cdot$ 12 h~\cite{zhidkov2018smartphone})} energy consumption on a Pixel 3 phone, and \underline{(iii)} 14.3\% error rate ($\downarrow$ 3.6\% than the state-of-the-art work of counting the number of speakers~\cite{stoter2018countnet}) under the low-resource setting of \textbf{8 hours ($\leq$ 8 hours~\cite{xu2020lrspeech})} training data.
\end{itemize}

\section{Related works}
\subsection{SAM based on counting concurrent speakers} 
Speech audio can provide important information about social ambiance, and can be applied to finding social hot spots~\cite{makhervaks2020combining}, event contexts~\cite{wang2014local}\cite{khan2015sensepresence}, and children's language environments~\cite{cristia2021thorough,ramirez2014look}. Specifically, \cite{wang2014local} measures social ambiance by inferring the occupancy and human chatter levels in local business scenarios; \cite{khan2015sensepresence} characterizes social ambiance from overlapping conversational data in a crowded environment. More recently, the number of concurrent speakers around an individual has been verified
to be associated with mental health symptoms~\cite{chen2021privacy}. The number of simultaneous speakers is leveraged as a proxy for the overall social activity in the associated environment based on the assumption that concurrent speakers provide fine-grained information of social ambiance~\cite{chen2021privacy}.   
 
Meanwhile, several recent studies~\cite{stoter2018countnet,peng2020competing} have shown that DNNs can achieve state-of-the-art accuracy in predicting the number of concurrent speakers as compared to non-DNN solutions. However, they either \underline{(i)} can only estimate up to five speakers, which cannot cover all social scenarios in daily life,  or \underline{(ii)} require more than 30 hours of training data, which can be challenging to acquire and label for SAM under clinical scenarios. In contrast, our proposed ERSAM can automatically deliver DNNs that perform energy-efficient
and real-time
SAM, based on a small-scale
training dataset which favors preserving privacy and timely feedback to fulfill the requirements of SAM under clinical scenarios. 

\subsection{Hardware-aware neural architecture search}
\label{sec:related_worke_hw_nas}
HW-NAS has been proposed with the aim of automating the search for efficient DNN structures under target hardware efficiency constraints (e.g., energy or latency on target devices) \cite{zhang2020dna}. Besides the early works based on reinforcement learning~\cite{tan2019mnasnet, howard2019searching}, ~\cite{wu2019fbnet,cai2018proxylessnas} develop differentiable HW-NAS following~\cite{liu2018darts} to significantly improve search efficiency. More recently,~\cite{cai2019once,yu2020bignas} propose to jointly train all sub-networks within the search space in a weight-sharing supernet and then locate the optimal architectures under different cost constraints without re-training or fine-tuning, thus reducing the cost of the whole search and training pipeline as compared to previous HW-NAS solutions. However, all these HW-NAS works are dedicated to computer vision or natural language processing applications and cannot be directly applied to SAM, due to the differences in input modalities, commonly used operators and dataset size. Note that although there exist prior works that apply NAS to speech recognition~\cite{chen2020darts,mo2020neural,kim2020evolved}, none of them target developing continuous, real-time, and on-device DNN-based SAM that meets all the requirements mentioned in Sec.~\ref{sec:intro}.

\subsection{Efficient knowledge distillation}
Although knowledge distillation~\cite{hinton2015distilling} is commonly used to boost the achievable accuracy vs. inference efficiency trade-offs of compact DNNs, it requires more training time due to the extra overhead of querying larger teacher models~\cite{gou2021knowledge,bommasani2021opportunities}.            

While there exist prior works that focus on efficient knowledge distillation, they either \underline{(i)} only focus on the data efficiency~\cite{wang2020neural,meng2019conditional} instead of the efficiency of the training latency or energy, or \underline{(ii)} are only designed for or verified on computer vision tasks with specially designed image operators~\cite{shen2021fast}. Specifically, ~\cite{meng2019conditional} leverages the insight that teacher models are not always perfect and bypasses the wrong predictions of teacher models to improve the accuracy of student models. Different from all the prior works mentioned above, our proposed efficient knowledge distillation scheme \underline{(i)} presents a different insight, \textcolor{black}{which is} the improved accuracy of the larger teacher model over the small student model is caused by the case when the smaller models are wrong and uncertain while the larger models are correct and certain and \underline{(ii)} targets compressing training/search cost.
\section{The proposed ERSAM framework}

\begin{figure}[b]
\vspace{-1em}
    \centering
    \includegraphics[width=1.0\linewidth]{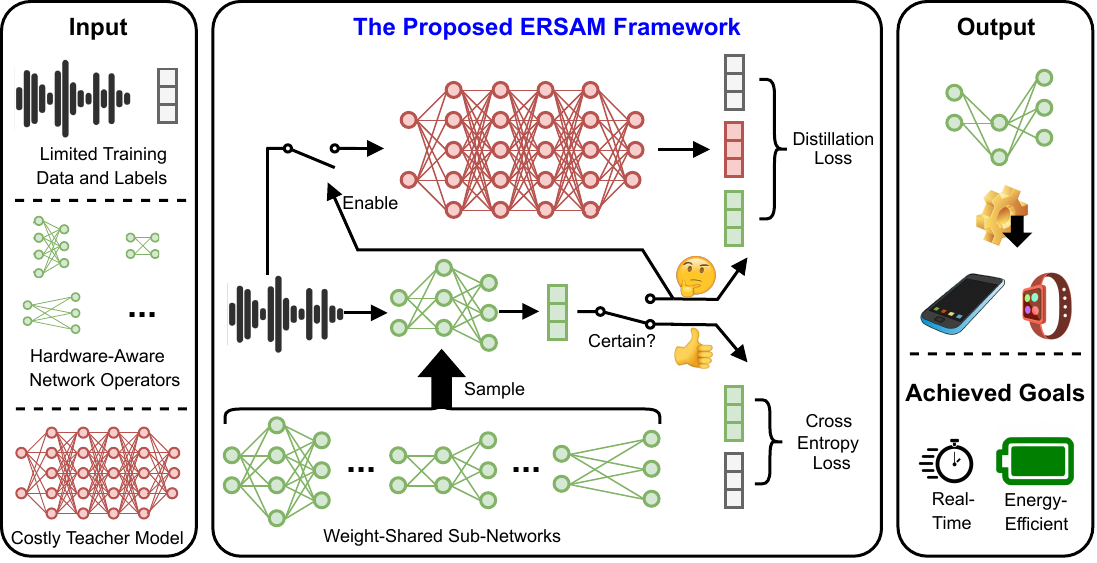}
    \vspace{-2em}
    \caption{Overview of our proposed ERSAM framework.}
    \label{fig:overview}
\end{figure}

\textbf{Overview.} 
As shown in Fig.~\ref{fig:overview}, our ERSAM accepts training data and corresponding labels of limited size, hardware-aware network operators, and a costly teacher model as inputs, and then automatically generates a dedicated DNN to enable continuous, real-time, on-device DNN based SAM. 
ERSAM integrates two enablers: \underline{(i)} a hardware-aware search space dedicated to SAM based on the cost profiling observations of state-of-the-art DNN based SAM models on mobile phones, and \underline{(ii)} an efficient knowledge distillation scheme to be embedded in the search and training process by only making queries when the student model is uncertain under the given input data. Specifically, during each iteration, we sample one sub-network from the search space, and then pass the training data to the sampled sub-network. If the measured uncertainty based on the sampled sub-network's predictions is higher than a pre-defined threshold, the query to the teacher model will be enabled, and the corresponding back-propagation will be performed to incorporate an extra distillation loss~\cite{hinton2015distilling}. Otherwise, the training will be the same as the standard training process without knowledge distillation. Note that ERSAM is built on top of a weight-sharing NAS ~\cite{cai2019once,yu2020bignas}, which trains all the sub-networks in a weight-sharing supernet and then locates the optimal one under different cost constraints without re-training; thus, the search cost is shared by training the supernet.

\begin{figure}[t]
    \centering
    \includegraphics[width=0.9\linewidth]{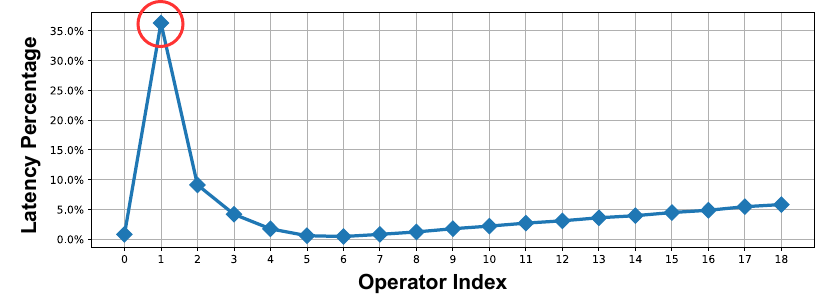}
    \vspace{-1.5em}
   \caption{Latency of different operators in wav2vec~\cite{schneider2019wav2vec} profiled on a Pixel 3~\cite{pixel3} phone.}
    \label{fig:wav2vec_profile}
    \vspace{-1em}
\end{figure}

\subsection{Enabler 1: Hardware-aware search space design}
\label{sec:space_design}

As pointed out in prior works, even for DNNs that have the same width, depth, or input resolution, the real hardware-cost (e.g., latency and/or energy) can be quite different on a target device~\cite{li2021hw,fu2022depthshrinker}. Thus, a hardware-aware search space is critical for achieving our target continuous, real-time, and on-device DNN-based SAM. Specifically, we first analyze the latency of each operator in wav2vec~\cite{schneider2019wav2vec}, a state-of-the-art speech recognition model. As shown in Fig.~\ref{fig:wav2vec_profile}, the bottleneck operator, marked with a red circle in Fig.~\ref{fig:wav2vec_profile}, has a cost that is $>$ 35\% of the whole model's latency on a Pixel 3 phone. We find that the above cost-dominant operator features both a large kernel size (i.e., 8) and input sequence length (i.e., $0.2 \times$ of the original input sequence length), while maintaining the same channel size as other operators (i.e., 512). As such, we design a hardware-aware search space for SAM that is abstracted from the model architecture of wav2vec~\cite{schneider2019wav2vec}, while avoiding cases where a large channel size, kernel size, and input sequence length simultaneously exist within the same operator. The developed search space in our ERSAM framework, inspired by our cost profiling observations on mobile phones, is summarized in Tab.~\ref{tab:search_space}.

\begin{table}[t]
\caption{Macro-architecture of our designed hardware-aware hardware search space for SAM.}
\vspace{-0.5em}
\centering
  \resizebox{0.95\linewidth}{!}
  {
    \begin{tabular}{c||cccc}
    \toprule
    Operator Type & \#Channels & \#Repeats & Kernel Size & Stride \\
    \midrule
    1D Conv. & 16 - 32 & \{1\} & 10 & 5 \\
    1D Conv. & 32 - 64  & \{1\}  & 8 & 4 \\
    1D Conv. & 64 - 128  & \{1, 2, 3\} & 4 & 2 \\
    1D Conv. & 128 - 256  & \{1, 2, 3\} & 1 & 1 \\
    1D Conv. & 128 - 256  & \{1, 2, 3\} & \{1, 2, 3\} & 1 \\
    1D Conv. & 128 - 256  & \{1, 2, 3\} & \{4, 5, 6\} & 1 \\
    1D Conv. & 128 - 256  & \{1, 2, 3\} & \{7, 8, 9\} & 1 \\
    1D Conv. & 128 - 256 & \{1, 2, 3\} & \{10, 11, 12\} & 1 \\
    \bottomrule
    \end{tabular}
    }
  \label{tab:search_space}
  \vspace{-1em}
\end{table}

\subsection{Enabler 2: Efficient knowledge distillation scheme}
\label{sec:efficient_kd}
The proposed efficient knowledge distillation scheme is based on our observation that the improved accuracy of larger models over smaller models for SAM is a result of the case where the smaller models are wrong and uncertain but the larger models are correct and certain. Specifically, we compare the correct and wrong predictions of a larger model called wav2vec~\cite{schneider2019wav2vec} and a smaller counterpart, a uniformly scaled wav2vec with only 6 (0.3$\times$) layers and 128 (0.25$\times$) channels, on the LibriSpeech-SAM dataset described in Sec.~\ref{sec:exp_settings}.

\begin{wrapfigure}{r}{0.54\linewidth}
    \centering 
    \vspace{-1.5em}
  \includegraphics[width=1.0\linewidth]{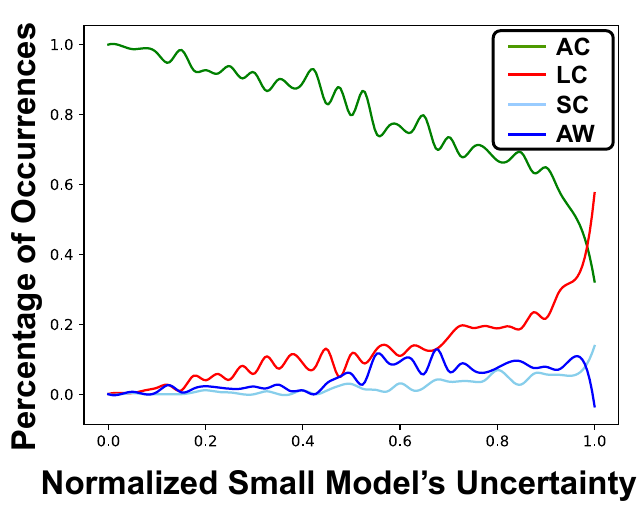}
  \vspace{-2em}
  \caption{The percentage of cases when varying the smaller model's uncertainties (AC: both correct, LC: only the larger model correct, SC: only the smaller model correct, and AW: both wrong).}
  \vspace{-1.5em}
\label{fig:big_vs_small}
\end{wrapfigure}

As shown in Fig.~\ref{fig:big_vs_small}, we can observe that the more uncertain the smaller model is, the more improvement can be gained by switching to the larger model from the smaller one, as evidenced by the sharp increase in the red curve corresponding to the case where only the large model is correct. Such an observation is also consistent with recent findings in natural language processing tasks~\cite{narayan2022predicting}, i.e., the larger models' advantages over smaller ones appear when the latter are not uncertain about the input data.
Leveraging the above observation, in the forward process of our proposed efficient knowledge distillation scheme, we first obtain the student model's classification probability distribution, and then compare it with the prior classification probability distribution to measure the uncertainty of the current input data, which is defined as follows:  

\vspace{-2.5em}
\begin{align}
    \begin{split} \label{eq:conv1}
    S_{sample} = - \log \left(\frac{1}{n}\sum^{n-1}_{i=0} (Y_i - \hat{Y_i})^2\right),
    \end{split} \\
    \begin{split} \label{eq:conv2}
    S_{batch} = \frac{1}{s}\sum^{s-1}_{j=0} S_{sample,j},
     \end{split}
\end{align}
\vspace{-1em}

\noindent where $S_{sample}$ and $S_{batch}$ denote the uncertainty score of one sample and a batch of samples, $s$ is the batch size, $Y_i$ is the output probability of the model for class $i$, $\hat{Y_i}$ is the prior probability of the labels for class $i$, and $n$ is the number of classes. To simplify the measurement, all our experiments are based on the assumption that different classes have the same prior probability, i.e., $\hat{Y_i} = \hat{Y_j}$ for any $i$ and $j$. As shown in Fig.~\ref{fig:overview}, if $S_{batch}$ is larger than the pre-defined threshold, then a query to the teacher model will be made; otherwise, this iteration will proceed with the standard training without knowledge distillation. The proposed efficient knowledge distillation scheme can achieve the same or even better accuracy compared with that of training with vanilla knowledge distillation~\cite{hinton2015distilling} yet at a similar training speed as the standard training, as verified in Sec.~\ref{sec:abl_efficient_kd}.
\section{Experiments results}
\subsection{Experiment settings}

\label{sec:exp_settings}
\textbf{Dataset.} To the best of our knowledge, there does not exist any realistic dataset labeled with the number of simultaneous speakers. Thus, we have synthesized a dataset with speech mixtures from LibriSpeech corpus\cite{panayotov2015librispeech}. To create a $k$-speaker mixture, where $k\in\{0, 1, 2, 3, 4, \geq5\}$, we first generate a 15-30 minutes recording by concatenating all the sentences from each speaker, and then randomly select an audio segment from each recording. Finally, segments from $k$ speakers are trimmed to 5 seconds and overlapped with each other to generate a speech mixture which is labeled as speaker count of $k$. Additionally, we include non-speech samples from the TUT Acoustic Scenes dataset \cite{Mesaros2018_IWAENC} in our training data when $k$ = 0 for scenarios where no speech is detected. Following the procedures above, 8 hours of speech mixtures are generated for training and validation. Meanwhile, the test subset with 2 hours of data is synthesized from a different set of speakers from LibriSpeech-test subset. The synthesized dataset is denoted as LibriSpeech-SAM and is used for all the experiments in this work. Specifically, to be fair with each possible speaker count, the dataset features balanced class distribution for both train and test subsets, i.e.,  $\sim1,000$ samples for each class in the training set and $\sim300$ samples in the testing set.

\textbf{Baselines and evaluation metrics.} We consider the following two models as our baselines: \underline{(i)} (uniformly scaled) wav2vec~\cite{schneider2019wav2vec}, one of the most common models for speech recognition. Because the original wav2vec~\cite{schneider2019wav2vec} exceeds the latency and energy budgets for being real-time on mobile devices, we scale it down by reducing its number of channels and number of layers uniformly to fulfill the requirements in Sec.~\ref{sec:intro}; \underline{(ii)} Countnet~\cite{stoter2018countnet}, the previous state-of-the-art for counting the number of speakers. For a fair comparison, we retrain it with the corresponding official implementation on our LibriSpeech-SAM dataset, because its accuracy reported in~\cite{stoter2018countnet} is trained on a different dataset. Moreover, we leverage the following evaluation metrics in all our experiments: \underline{(i)} the \textbf{error rate} on the test set of LibriSpeech-SAM; \underline{(ii)} the \textbf{latency} of an audio segment of 5 seconds on a Pixel 3 Phone; \underline{(iii)} the \textbf{energy consumption} on a Pixel 3 Phone when running all day (i.e., 12 h active time). 

\textbf{Experiment platforms.} All the training and search processes are performed on a workstation with a commonly-used NVIDIA 2080 Ti GPU to match the settings of the training on the local edge (e.g., on personal desktops or laptops). The latency and energy consumption of DNN's inference are measured on a Pixel 3 Phone by running the model using the official tflite benchmark binary~\cite{pixel3_latency_measurement} and being monitored by the Qualcomm Snapdragon Profiler~\cite{snapdragon_profiler}.

\begin{table}[b]
\vspace{-1.3em}
\caption{Compare the DNN searched by our proposed ERSAM with the previous state-of-the-art.}
\centering
\vspace{-0.5em}
  \resizebox{\linewidth}{!}
  {
    \begin{tabular}{c||ccc}
    \toprule
    \multirow{2}{*}{\textbf{Method}} & \textbf{Error Rate on} & \textbf{Latency on} & \textbf{Energy consumption}  \\
     & \textbf{LibriSpeech-SAM (\%)} &  \textbf{Pixel 3 (ms)} &  \textbf{on Pixel 3 (mW$\cdot$h)} \\
     \midrule
     Countnet~\cite{stoter2018countnet} & 16.7 & 492  & 422 \\
     (Uniformly Scaled) wav2vec~\cite{schneider2019wav2vec} & 17.4 & 50 & 40 \\
     \textbf{ERSAM} & \textbf{14.3} & \textbf{47} & \textbf{38}  \\
    \bottomrule
    \end{tabular}
    }
  \label{tab:compare_with_sota}
\end{table}

\subsection{Compare with state-of-the-art}
We summarize the comparison of our proposed ERSAM with the previous state-of-the-art in both Tab.~\ref{tab:compare_with_sota} and Fig.~\ref{fig:performance_overview}. We can observe that ours can achieve the best error rate vs. hardware cost (e.g., latency or energy consumption) trade-offs, verifying our ERSAM's ability to deliver continuous, real-time, and on-device DNN-based SAM in an automatic manner without manually searching.

\subsection{Ablation study on the proposed efficient knowledge distillation scheme}

To better understand the efficacy and efficiency of our proposed efficient knowledge distillation scheme introduced in Sec.~\ref{sec:efficient_kd}, we compare it with the vanilla knowledge distillation~\cite{hinton2015distilling} and the standard training without any knowledge distillation in terms of the error rate of the finally delivered DNN and the corresponding search and training time. 
As summarized in Tab.~\ref{tab:abs_efficient_distillation}, we can observe that \underline{(i)} the vanilla knowledge distillation can achieve a lower error rate (e.g., $\downarrow$ 2.4\%) but 
with $2.95 \times$ of the search and training cost, as compared to standard training (i.e., ERSAM w/o KD); \underline{(ii)} our proposed efficient knowledge distillation achieves the lowest error rate among all competitors while maintaining a similar (e.g., $1.05 \times$) search and training cost with standard training. Such observations imply that our proposed efficient knowledge distillation scheme can be used as a plug-and-play module in HW-NAS, thanks to its advantage of achieving a lower error rate than vanilla knowledge distillation without requiring extra search and training costs.

\label{sec:abl_efficient_kd}
\begin{table}[h]
\vspace{-0.5em}
\caption{Comparison among ERSAM w/o knowledge distillation (KD), w/ vanilla KD~\cite{hinton2015distilling}, and w/ our proposed efficient KD.}
\centering
\vspace{-0.5em}
  \resizebox{0.9\linewidth}{!}
  {
    \begin{tabular}{c||cc}
    \toprule
    \multirow{2}{*}{\textbf{Method}} & \textbf{Error Rate on } & \textbf{Search Cost}  \\
     & \textbf{LibriSpeech-SAM (\%)} & \textbf{(GPU hours)}  \\
     \midrule
     ERSAM w/o KD & 16.9   & 2.1 \\
     ERSAM w/ vanilla KD~\cite{hinton2015distilling} & 14.5 ($\downarrow$ 2.4)   & 6.2 (2.95 $\times$) \\
     \textbf{ERSAM w/ our proposed KD} &  \textbf{ 14.3 ($\downarrow$ 2.6) }  & \textbf{2.2 (1.05 $\times$)} \\
    \bottomrule
    \end{tabular}
    }
  \label{tab:abs_efficient_distillation}
  \vspace{-1.5em}
\end{table}

\subsection{Generalize to real-world scenarios}

To simulate speech audio collected from realistic scenarios, we \underline{(i)} apply MUSAN noises \cite{snyder2015musan} (12-18 dB) to each recording, \underline{(ii)} include domain shifts caused by recording devices using LibriAdapt~\cite{mathur2020libri}, and \underline{(iii)} construct a few noisy SAM datasets following Sec.~\ref{sec:exp_settings}. As summarized in Tab.~\ref{tab:generalization}, we can observe that the proposed ERSAM can still achieve a low error rate under challenging settings and beat the uniformly scaled wav2vec~\cite{baevski2020wav2vec} in terms of achieved error rate vs. hardware cost trade-offs (i.e., a $\downarrow$ 2.7\% $\sim$ 5.0\% lower error rate under similar latency and energy consumption).

\begin{table}[h]
\vspace{-0.5em}
\caption{Generalize the DNN searched by our proposed ERSAM and uniformly scaled wav2vec~\cite{baevski2020wav2vec} to the dataset with noises and device diversity for simulating the scenarios of real-world application.}
\centering
\vspace{-0.5em}
  \resizebox{\linewidth}{!}
  {
    \begin{tabular}{c||cc}
    \toprule
    Error Rate (\%) & (scaled) wav2vec~\cite{schneider2019wav2vec} & \textbf{ERSAM} \\
     \midrule
     LibriSpeech-SAM & 17.4 & \textbf{14.3} ($\downarrow$ 3.1) \\
     LibriSpeech-SAM, Noisy & 20.7 & \textbf{17.1} ($\downarrow$ 3.6) \\
     LibriAdapt-SAM on Pseye & 30.4 & \textbf{27.5} ($\downarrow$ 2.9) \\
     LibriAdapt-SAM on Respeaker & 28.6 & \textbf{23.6} ($\downarrow$ 5.0) \\
     LibriAdapt-SAM on Shure & 28.7 & \textbf{25.9} ($\downarrow$ 2.8) \\
    \bottomrule
    \end{tabular}
    }
  \label{tab:generalization}
  \vspace{-1.5em}
\end{table}
\section{Conclusion}
In this work, we propose ERSAM, a dedicated framework for developing continuous, real-time, on-device DNN based SAM to meet all three requirements on the execution latency (e.g., $\leq$ 5 seconds), energy (e.g., $\leq$ 700mW $\cdot$ 12 h), and size of training data  (e.g., audio recordings of $\leq$ 8 hours in total) towards practical adoption. This framework consists of: \underline{(i)} a hardware-aware search space dedicated to SAM; and \underline{(ii)} an efficient knowledge distillation scheme to be embedded in the search and training processes. The DNN model \textcolor{black}{searched by our ERSAM} achieves better accuracy vs. hardware efficiency trade-offs than previous state-of-the-art works and fulfills all the requirements towards continuous, real-time, on-device SAM.

\bibliographystyle{IEEEbib}

\bibliography{refs}

\begin{thebibliography}{10}

\bibitem{chen2021privacy}
Wenwan Chen et~al.,
\newblock ``Privacy-preserving social ambiance measure from free-living speech
  associate with chronic depressive and psychotic disorders,''
\newblock {\em Frontiers in psychiatry}, 2021.

\bibitem{wang2014local}
He~Wang et~al.,
\newblock ``Local business ambience characterization through mobile audio
  sensing,''
\newblock in {\em Proceedings of the 23rd international conference on World
  wide web}, 2014.

\bibitem{baevski2020wav2vec}
Alexei Baevski et~al.,
\newblock ``wav2vec 2.0: A framework for self-supervised learning of speech
  representations,''
\newblock {\em Advances in Neural Information Processing Systems}, vol. 33,
  2020.

\bibitem{pixel3}
{Google LLC.},
\newblock ``{Pixel 3},'' \url{https://g.co/kgs/pVRc1Y}, accessed 2020-09-01.

\bibitem{pytorch_mobile_wav2vec2}
{Meta Inc.},
\newblock ``{Speech Recognition on Android with Wav2Vec2 },''
  \url{https://github.com/pytorch/android-demo-app/blob/master/SpeechRecognition},
  accessed 2021-09-01.

\bibitem{zhidkov2018smartphone}
Sergey Zhidkov et~al.,
\newblock ``On smartphone power consumption in acoustic environment monitoring
  applications,''
\newblock {\em Applied System Innovation}, vol. 1, no. 1, 2018.

\bibitem{xu2020lrspeech}
Jin Xu et~al.,
\newblock ``Lrspeech: Extremely low-resource speech synthesis and
  recognition,''
\newblock in {\em Proceedings of the 26th ACM SIGKDD}, 2020.

\bibitem{wu2019fbnet}
Bichen Wu et~al.,
\newblock ``Fbnet: Hardware-aware efficient convnet design via differentiable
  neural architecture search,''
\newblock in {\em Proceedings of the IEEE CVPR}, 2019.

\bibitem{cai2019once}
Han Cai et~al.,
\newblock ``Once-for-all: Train one network and specialize it for efficient
  deployment,''
\newblock {\em arXiv:1908.09791}, 2019.

\bibitem{yu2020bignas}
Jiahui Yu et~al.,
\newblock ``Bignas: Scaling up neural architecture search with big single-stage
  models,''
\newblock in {\em ECCV}. Springer, 2020.

\bibitem{hinton2015distilling}
Geoffrey Hinton et~al.,
\newblock ``Distilling the knowledge in a neural network,''
\newblock {\em arXiv:1503.02531}, 2015.

\bibitem{gou2021knowledge}
Jianping Gou et~al.,
\newblock ``Knowledge distillation: A survey,''
\newblock {\em International Journal of Computer Vision}, vol. 129, no. 6,
  2021.

\bibitem{deng2009imagenet}
Jia Deng et~al.,
\newblock ``Imagenet: A large-scale hierarchical image database,''
\newblock in {\em 2009 IEEE Conference on CVPR}, 2009.

\bibitem{stoter2018countnet}
Fabian-Robert St{\"o}ter et~al.,
\newblock ``Countnet: Estimating the number of concurrent speakers using
  supervised learning,''
\newblock {\em IEEE/ACM TASLP}, vol. 27, no. 2, 2018.

\bibitem{makhervaks2020combining}
Dave Makhervaks et~al.,
\newblock ``Combining acoustics, content and interaction features to find hot
  spots in meetings,''
\newblock in {\em ICASSP 2020-2020 IEEE ICASSP}. IEEE, 2020, pp. 8054--8058.

\bibitem{khan2015sensepresence}
Md~Abdullah Al~Hafiz Khan et~al.,
\newblock ``Sensepresence: Infrastructure-less occupancy detection for
  opportunistic sensing applications,''
\newblock in {\em 2015 16th IEEE MDM}, 2015.

\bibitem{cristia2021thorough}
Alejandrina Cristia et~al.,
\newblock ``A thorough evaluation of the language environment analysis (lena)
  system,''
\newblock {\em Behavior Research Methods}, vol. 53, no. 2, pp. 467--486, 2021.

\bibitem{ramirez2014look}
Nair{\'a}n Ram{\'\i}rez-Esparza et~al.,
\newblock ``Look who's talking: Speech style and social context in language
  input to infants are linked to concurrent and future speech development,''
\newblock {\em Developmental science}, vol. 17, no. 6, pp. 880--891, 2014.

\bibitem{peng2020competing}
Chao Peng et~al.,
\newblock ``Competing speaker count estimation on the fusion of the spectral
  and spatial embedding space.,''
\newblock in {\em INTERSPEECH}, 2020.

\bibitem{zhang2020dna}
Yongan Zhang et~al.,
\newblock ``Dna: Differentiable network-accelerator co-search,'' 2020.

\bibitem{tan2019mnasnet}
Mingxing Tan et~al.,
\newblock ``Mnasnet: Platform-aware neural architecture search for mobile,''
\newblock in {\em Proceedings of the IEEE Conference on Computer Vision and
  Pattern Recognition}, 2019.

\bibitem{howard2019searching}
Andrew Howard et~al.,
\newblock ``Searching for mobilenetv3,''
\newblock in {\em Proceedings of the IEEE ICCV}, 2019.

\bibitem{cai2018proxylessnas}
Han Cai et~al.,
\newblock ``Proxylessnas: Direct neural architecture search on target task and
  hardware,''
\newblock {\em arXiv:1812.00332}, 2018.

\bibitem{liu2018darts}
Hanxiao Liu et~al.,
\newblock ``Darts: Differentiable architecture search,''
\newblock {\em arXiv:1806.09055}, 2018.

\bibitem{chen2020darts}
Yi-Chen Chen et~al.,
\newblock ``Darts-asr: Differentiable architecture search for multilingual
  speech recognition and adaptation,''
\newblock {\em arXiv:2005.07029}, 2020.

\bibitem{mo2020neural}
Tong Mo et~al.,
\newblock ``Neural architecture search for keyword spotting,''
\newblock {\em arXiv:2009.00165}, 2020.

\bibitem{kim2020evolved}
Jihwan Kim et~al.,
\newblock ``Evolved speech-transformer: Applying neural architecture search to
  end-to-end automatic speech recognition.,''
\newblock in {\em INTERSPEECH}, 2020.

\bibitem{bommasani2021opportunities}
Rishi Bommasani et~al.,
\newblock ``On the opportunities and risks of foundation models,''
\newblock {\em arXiv:2108.07258}, 2021.

\bibitem{wang2020neural}
Dongdong Wang et~al.,
\newblock ``Neural networks are more productive teachers than human raters:
  Active mixup for data-efficient knowledge distillation from a blackbox
  model,''
\newblock in {\em Proceedings of the IEEE CVPR}, 2020.

\bibitem{meng2019conditional}
Zhong Meng, Jinyu Li, Yong Zhao, and Yifan Gong,
\newblock ``Conditional teacher-student learning,''
\newblock in {\em ICASSP 2019-2019 IEEE International Conference on Acoustics,
  Speech and Signal Processing (ICASSP)}. IEEE, 2019, pp. 6445--6449.

\bibitem{shen2021fast}
Zhiqiang Shen et~al.,
\newblock ``A fast knowledge distillation framework for visual recognition,''
\newblock {\em arXiv:2112.01528}, 2021.

\bibitem{schneider2019wav2vec}
Steffen Schneider et~al.,
\newblock ``wav2vec: Unsupervised pre-training for speech recognition,''
\newblock {\em arXiv:1904.05862}, 2019.

\bibitem{li2021hw}
Chaojian Li et~al.,
\newblock ``Hw-nas-bench: Hardware-aware neural architecture search
  benchmark,''
\newblock {\em arXiv:2103.10584}, 2021.

\bibitem{fu2022depthshrinker}
Yonggan Fu et~al.,
\newblock ``Depthshrinker: A new compression paradigm towards boosting
  real-hardware efficiency of compact neural networks,''
\newblock {\em arXiv:2206.00843}, 2022.

\bibitem{narayan2022predicting}
Taman Narayan et~al.,
\newblock ``Predicting on the edge: Identifying where a larger model does
  better,''
\newblock {\em arXiv:2202.07652}, 2022.

\bibitem{panayotov2015librispeech}
Vassil Panayotov et~al.,
\newblock ``Librispeech: an asr corpus based on public domain audio books,''
\newblock 2015.

\bibitem{Mesaros2018_IWAENC}
Annamaria Mesaros et~al.,
\newblock ``Acoustic scene classification: An overview of {DCASE} 2017
  challenge entries,''
\newblock in {\em 2018 16th IWAENC}, September 2018.

\bibitem{pixel3_latency_measurement}
{Google LLC.},
\newblock ``{Performance measurement},''
  \url{https://www.tensorflow.org/lite/performance/measurement}, accessed
  2021-05-21.

\bibitem{snapdragon_profiler}
{Qualcomm Technologies, Inc.},
\newblock ``{Snapdragon Profiler},''
  \url{https://developer.qualcomm.com/software/snapdragon-profiler}, accessed
  2022-03-21.

\bibitem{snyder2015musan}
David Snyder et~al.,
\newblock ``Musan: A music, speech, and noise corpus,''
\newblock {\em arXiv:1510.08484}, 2015.

\bibitem{mathur2020libri}
Akhil et~al Mathur,
\newblock ``Libri-adapt: a new speech dataset for unsupervised domain
  adaptation,''
\newblock in {\em ICASSP 2020-2020 ICASSP}. IEEE, 2020, pp. 7439--7443.

\end{thebibliography}

\end{document}